\documentclass[runningheads]{llncs}

\usepackage{eccv}

\usepackage{eccvabbrv}

\usepackage{graphicx}
\usepackage{booktabs}

\usepackage{algorithm}
\usepackage{algorithmic}
\usepackage{multirow}
\usepackage{comment}

\usepackage[accsupp]{axessibility}

\usepackage{hyperref}

\usepackage{orcidlink}

\usepackage{enumitem}
\usepackage{microtype}
\usepackage{makecell}
\newcommand{\method}{CACFM\xspace}

\begin{document}

\title{Curvature-Adaptive Consistency Flow Matching: Autonomous Trajectory Optimization via Reinforcement Learning} 

\titlerunning{Curvature-Adaptive Consistency Flow Matching}

\author{Songtao Tian\inst{1}\thanks{Equal contribution. \quad$^\dag$\,Corresponding author.} \and
Guhan Chen\inst{1}$^{\star}$ \and
Bohan Li\inst{1} \and Jingyi Ma\inst{1} \and Zixiong Yu\inst{2,1}$^\dag$}

\authorrunning{S. Tian et al.}

\institute{\textsuperscript{1}Tsinghua University\quad
\textsuperscript{2}Huawei Large Model Data Technology Lab\\ 
\email{tiansongtao.2020@tsinghua.org.cn}\quad\email{chen-gh23@mails.tsinghua.edu.cn}\quad\email{\{libh19,majy20,yuzx19\}@tsinghua.org.cn}}

\maketitle

\begin{abstract}

Consistency distillation has significantly accelerated diffusion-model inference, but its sampling dynamics remain underexplored. We reveal an asymmetry: although Logit-Normal sampling priors work well for standard iterative generation, consistency distillation exhibits a different difficulty profile (\eg, U-shaped), with optimization bottlenecks concentrated at the boundary stages rather than intermediate steps. To address the limitations of static sampling under evolving learning demands, we propose Curvature-Adaptive Consistency Flow Matching (\method). By formulating distillation as a dynamic decision process, \method uses a lightweight reinforcement learning agent to probe Probability Flow ODE trajectories and construct an efficiency-oriented curriculum that prioritizes critical regions without manual scheduling. Combined with Flow-adapted DMD and adversarial consistency objectives, our RL-based scheduler achieves state-of-the-art results on large-scale models such as FLUX and SDXL, mitigating structural deformities and preserving high-frequency details in extreme few-step regimes.
\end{abstract}

\begin{figure*}[t]
 \centering
 \includegraphics[width=0.8\linewidth]{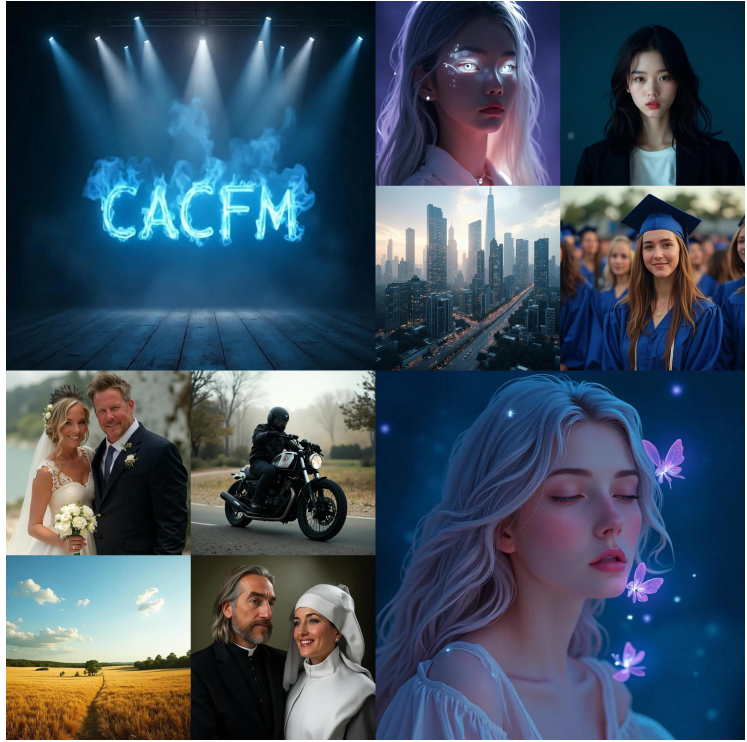}
 \caption{1024$\times$1024 samples produced by our 4-step generator distilled from FLUX.1-dev.}
 \label{fig:FLUX_result}
\end{figure*}

\section{Introduction}\label{sec:intro}
In recent years, artificial intelligence (AI) systems have become a central pillar of modern computation. Although recent theoretical studies have advanced our understanding of these systems \cite{lai2023generalizationabilitywideresidual,li2024eigenvalue}, this understanding remains incomplete \cite{chen2024impacts,yu2025divergence}. Modern AI models have demonstrated strong capabilities in language understanding \cite{yu-etal-2026-mathagent,vaswani2017attention}, visual synthesis \cite{lin20253d,liang2026render}, video generation \cite{Li2026ReTrack,Hu2026REFINE,he2024Prior}, and multimodal reasoning \cite{zhang2025vision,yu2024samWav2lip}. These advances have been driven by progress in architectures \cite{he2016deep,yu2026branch}, data construction \cite{He_2026_CVPR_CineMatte,rao2025data,xiao2026points}, and algorithm design \cite{shao2024deepseekmathpushinglimitsmathematical,wu2026toolaugmentedpolicyoptimizationsynergizing,zhao2026advances}.

Within generative modeling, diffusion models and their recent extension to Flow Matching (FM) frameworks have emerged as a powerful paradigm across multiple domains \cite{sohl2015deep,song2020score,lipman2022flow}. Notably, modern large-scale transformer-based architectures \cite{vaswani2017attention}, such as FLUX \cite{flux2024} and SD3 \cite{esser2024scaling} utilize the Flow Matching formulation, which relies on deterministic Probability Flow ODEs (PF-ODEs) to transform noise into data \cite{lipman2022flow}. While this formulation provides flexible control over the trade-off between computational cost and output quality, it inherently requires numerous sequential evaluations of computationally intensive neural networks, creating a significant bottleneck for real-time applications.

To mitigate this computational burden, extensive research has focused on acceleration techniques, ranging from advanced ODE solvers \cite{karras2022elucidating,lu2022dpm} to distillation-based approaches like Consistency Models (CMs \cite{song2023consistency}) and Phased Consistency Models (PCMs \cite{wang2024phased}). However, existing methods exhibit a distinct static nature in their sampling strategies: they predominantly rely on pre-defined discretization schemes or heuristic priors, such as basic uniform sampling or directly adopting the Logit-Normal distribution \cite{esser2024scaling} widely utilized in Rectified Flow training.

However, we contend that high-dimensional generative flows in consistency distillation exhibit pronounced Geometric Heterogeneity, where curvature and optimization difficulty vary drastically across timesteps. A uniform allocation of computational budget fails to resolve high-curvature regions. Furthermore, directly adopting static heuristics like the Logit-Normal distribution lacks the adaptive flexibility to accommodate the unique optimization dynamics of consistency distillation, especially as optimization focal points shift during training.

To validate this perspective, we compute the Oracle Consistency Error along a converged teacher trajectory to reveal the true difficulty distribution. Building on this, we propose Curvature-Adaptive Consistency Flow Matching (\method; see Fig.~\ref{fig:FLUX_result} for visual examples), which reformulates consistency distillation as a dynamic decision process. Instead of manually designing sampling schedules, we deploy a lightweight Reinforcement Learning (RL) agent to dynamically probe the trajectory, identifying and prioritizing sub-trajectories with higher optimization returns. This leads to the following fundamental discovery:

\vspace{-5mm}
\begin{figure*}[h]
\begin{minipage}[t]{0.5\textwidth} 
\vspace{0pt}
\normalsize
\begin{enumerate}[leftmargin=*,nosep]
    \item \textbf{U-Shaped Difficulty Profile:} 
    Contrary to the common observation in standard iterative generation training that intermediate steps are the most challenging (modeled by Logit-Normal sampling), the consistency error analysis reveals that in consistency distillation tasks, the boundary stages are more challenging than the intermediate phases (Fig.~\ref{fig:learn_shape}).
    \item \textbf{Emergent Curriculum Learning:} 
    Beyond learning the aforementioned curve, 
    the agent develops a coarse-to-fine strategy, progressing from establishing global structures in early training to refining high-frequency details in later stages (Fig.~\ref{fig:rl_analysis}).
\end{enumerate}
\end{minipage}
\hfill
\begin{minipage}[t]{0.47\textwidth} 
\vspace{7pt}
\centering
\includegraphics[width=\linewidth]{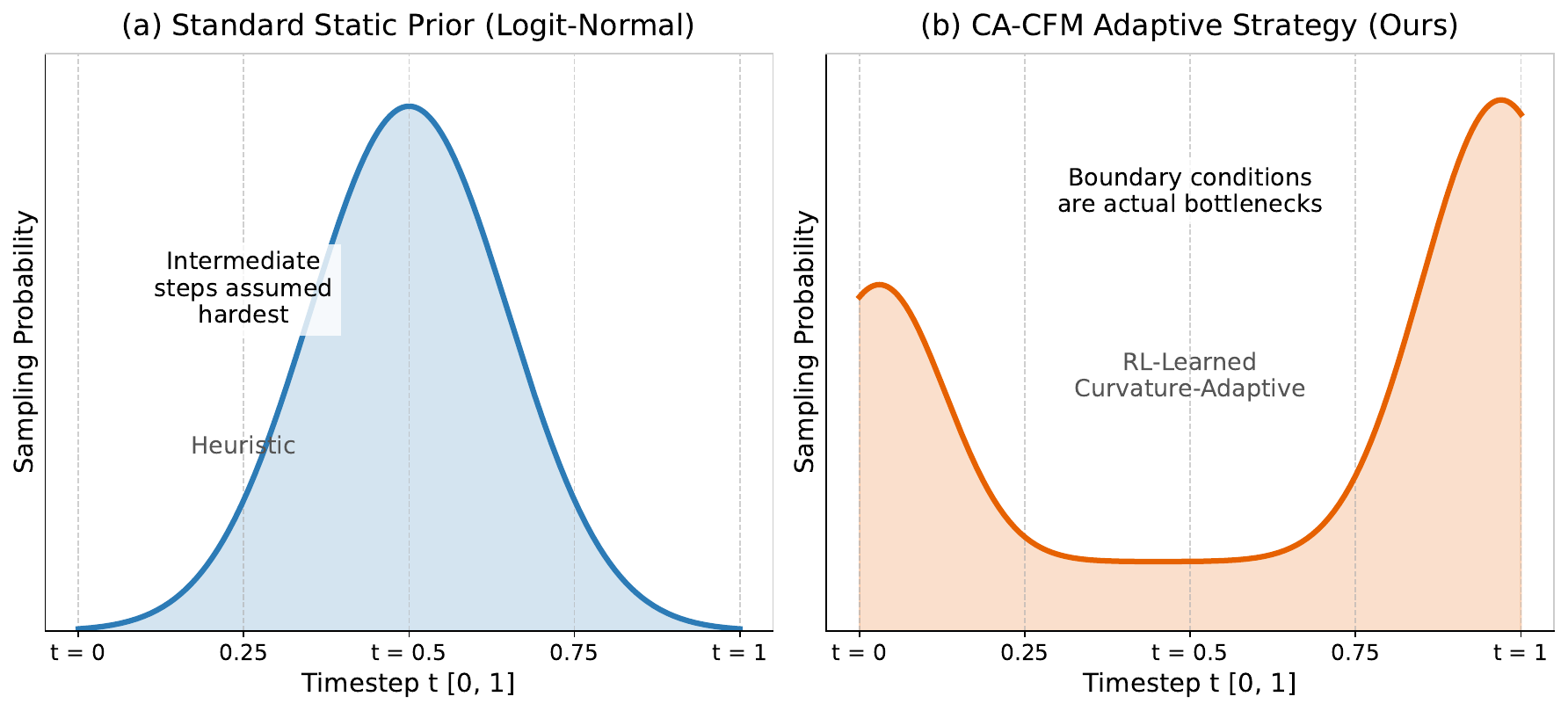}
\captionof{figure}{\textbf{Comparing sampling distributions along the FM trajectory.} (a) The Logit-Normal prior assumes that intermediate steps are the most important. (b) The U-shaped distribution identifies boundary stages (initialization and final refinement) as the primary bottlenecks, and our agent learns this pattern (see Fig.~\ref{fig:rl_analysis} for experimental results).}
\label{fig:learn_shape}
\end{minipage}
\vspace{-6.6mm}
\end{figure*}

While curvature-adaptive sampling effectively allocates the training budget to resolve trajectory bottlenecks, extreme few-step generation still suffers from distributional shift at the manifold level. To further enhance fidelity, we integrate Flow-adapted Distribution Matching Distillation (DMD), inspired by \cite{yin2024one}, and adversarial consistency as complementary training objectives; the core algorithmic novelty of \method remains the RL-based temporal scheduler. Extensive experiments on FLUX and SDXL architectures demonstrate that \method establishes new state-of-the-art (SOTA) performance. By intelligently allocating the computational budget to regions of high geometric complexity, our method achieves superior FID and aesthetic quality, proving that \textit{learning where to look} is as important as \textit{learning how to match}. 
Our contributions are threefold:
\begin{itemize}[leftmargin=*,nosep]
    \item \textbf{Discovery of Geometric Bottlenecks:} We reveal that consistency distillation exhibits a boundary-dominated (\eg, U-shaped) difficulty profile, contrary to the intermediate-focused priors of standard iterative generation.
    \item \textbf{Curvature-Adaptive Optimization via RL:} We propose \method, the first framework that dynamically adapts to the geometric heterogeneity of Flow Matching trajectories, replacing static heuristics with an agent. 
    \item \textbf{Adaptive Scheduler with Complementary Distillation Losses:}
    \method leverages an RL scheduler and complementary distillation losses, outperforming existing acceleration methods on large-scale benchmarks. 
\end{itemize}

\section{Related Work}
\subsubsection{Flow Matching and Static Consistency Distillation. }
FM \cite{liu2022flow,albergo2022building,lipman2022flow} has become a prominent framework for continuous-time generative modeling, using PF-ODEs to transform noise into data. Recent architectures such as FLUX \cite{flux2024} and SD3 \cite{esser2024scaling} achieve remarkable quality but require expensive iterative sampling. 
To accelerate inference, CMs \cite{song2023consistency} and Latent CMs (LCMs \cite{luo2023latent}) map trajectories directly to solutions, while PCMs \cite{wang2024phased} split trajectories into sub-segments.

\textit{The Static Prior Limitation:} Existing distillation methods rely on static sampling priors, from uniform sampling in CMs to Logit-Normal distributions \cite{esser2024scaling}, and thus cannot adapt to varying flow curvature. While adaptive sampling has been explored for data-level selection in broader training contexts \cite{rao-etal-2026-dynamic,yao2024swift}, \method extends this idea to trajectory-level policy learning over PF-ODE sub-trajectories.

\subsubsection{Efficient Distillation. }
Recent works improve few-step generation from complementary angles. SANA-Sprint \cite{chen2025sanasprint} combines consistency distillation with architectural and adversarial stabilizers, while ADM \cite{lu2025adversarial} uses adversarial distribution matching for image and video synthesis. Beyond diffusion, efficient student training has also been explored across diverse distillation settings \cite{rao2024pesfkd,li2025frequency,li2025ammkd}.

\textit{Teacher Matching vs. Trajectory Allocation:} These methods mainly improve \textit{how to match} the teacher distribution. In contrast, \method focuses on \textit{where to learn} along the trajectory, making our scheduler orthogonal to these loss-, teacher-, and architecture-centric advances.

\subsubsection{RL in Generative Models. }
RL has been widely applied to diffusion models, mainly for downstream alignment. Methods such as DDPO \cite{wallace2023diffusion} and DPOK \cite{fan2023dpok} frame the denoising process as a Markov Decision Process to optimize objectives such as human preference scores (HPS) or image compressibility. More broadly, feedback-driven policy learning has also been explored for adaptive solver control and rendering-aware generation outside diffusion distillation \cite{mai2026neural,liang2026vanim}.

\textit{Structural vs. Alignment Optimization:} 
Unlike these approaches, which use RL to optimize generation objectives (what to generate), \method employs RL to optimize training structure (how to learn). We treat sub-trajectory selection as a decision process to discover geometric bottlenecks in the flow.

\section{Preliminaries}
\subsubsection{FM and Geometric Complexity.}
FM models define a continuous-time probability path $p_t(\mathbf{x})$ that transports the data distribution $p_0(\mathbf{x})$ to a simple prior $p_1(\mathbf{x}) = \mathcal{N}(0,\mathbf{I})$. Unlike standard diffusion formulations based on stochastic differential equations, FM is defined through ODE dynamics: \(\mathrm{d} \mathbf{x}_t/\mathrm{d} t = \boldsymbol{v}(\mathbf{x}_t, t),~t \in [0,1]\), where $\boldsymbol{v}: \mathbb{R}^d \times [0,1] \to \mathbb{R}^d$ is a time-dependent vector field that induces the flow map $\phi_t(\mathbf{x})$, transporting data to noise during the forward process.

Specifically within the context of Optimal Transport Flow Matching, the forward process is defined by a linear interpolation between a target data sample $\mathbf{x}_0 \sim p_0$ and standard Gaussian noise $\mathbf{x}_1 \sim p_1$: \(\mathbf{x}_t = (1-t)\mathbf{x}_0 + t\mathbf{x}_1\). This conditional path induces a constant conditional vector field $\boldsymbol{u}_t(\mathbf{x}|\mathbf{x}_0, \mathbf{x}_1) = \mathbf{x}_1 - \mathbf{x}_0$. The objective of FM is to train a neural network $\boldsymbol{v}_\phi(\mathbf{x}, t)$ to regress this target vector field by minimizing the expected mean squared error:
\[
\mathcal{L}_{\text{FM}}(\phi) = \mathbb{E}_{t, \mathbf{x}_0, \mathbf{x}_1} \left[ \| \boldsymbol{v}_\phi(\mathbf{x}_t, t) - (\mathbf{x}_1 - \mathbf{x}_0) \|^2 \right].
\]

Once trained, the generation (sampling) process involves solving the learned ODE backward from an initial noise sample $\mathbf{x}_1 \sim \mathcal{N}(0,\mathbf{I})$ to $t=0$. The learned vector field $\boldsymbol{v}_{\phi}$ often exhibits geometric heterogeneity: it varies rapidly in high-curvature regions of the generative trajectory while remaining smooth in others. Consequently, accurately solving the learned ODE using standard numerical solvers requires many evaluations to traverse these geometrically complex regions.

\subsubsection{Consistency Distillation and the Sampling Bias.}
CMs \cite{song2023consistency} accelerate sampling by learning to map any point $\mathbf{x}_t$ on the PF-ODE trajectory directly to its origin. To mitigate the difficulty of learning long-range mappings, PCMs \cite{wang2024phased} partition the trajectory into $M$ sub-segments using boundaries $\{s_m\}_{m=0}^M$. 

Ideally, a model learns a local consistency function $\boldsymbol{f}^m_{\boldsymbol{\theta}}$ for each sub-trajectory $[s_m, s_{m+1}]$. The training objective utilizes a discretized grid $\{t_n\}_{n=0}^N$ and a one-step ODE solver estimate $\hat{\mathbf{x}}^{\boldsymbol{\phi}}_{t_n}$, formulated as:
\begin{equation}\label{eq:pcm_loss}
\mathcal{L}^{\text{PCM}}(\boldsymbol{\theta};\boldsymbol{\phi}) = 
\mathbb{E}\big[\lambda(t_n)\,d\big(\boldsymbol{f}^m_{\boldsymbol{\theta}}(\mathbf{x}_{t_{n+1}},t_{n+1}),\boldsymbol{f}^m_{\boldsymbol{\theta}^-}(\hat{\mathbf{x}}^{\boldsymbol{\phi}}_{t_n},t_n)\big)\big] 
\end{equation}
where $d(\cdot, \cdot)$ denotes a distance metric, expectations are taken over $\mathbb{P}(m)$, $\mathbb{P}(n|m)$ and $\mathbb{P}({\mathbf x_{t_{n+1}}|n,m})$, \ie, uniform distributions of sub-trajectory indices $m$, interval indices $n$, and distribution $\mathbb{P}_{t_{n+1}}$. The term $\boldsymbol{\theta}^-$ denotes the exponential moving average of the parameters, updated via $\boldsymbol{\theta}^- \leftarrow \mu \boldsymbol{\theta}^- + (1-\mu)\boldsymbol{\theta}$.

The key bottleneck in Eq.~\eqref{eq:pcm_loss} is the sampling distribution $\mathbb{P}(m)$. Existing methods sample the sub-trajectory index $m$ either uniformly or via a fixed Logit-Normal distribution, implicitly assuming that consistency difficulty is either homogeneous (Uniform) or concentrated in the middle (Logit-Normal). 
We argue that this static assumption is flawed. As we will show in Fig.~\ref{fig:rl_analysis}, the true "consistency bottlenecks" are often located at the boundaries (\eg, U-shaped).

\section{Methodology: Curvature-Adaptive Optimization}
The central hypothesis of this work is that the PF-ODE exhibits non-uniform geometric complexity: boundary regions require precise structural alignment, and optimization bottlenecks shift dynamically throughout distillation.

Static sampling strategies (Uniform, Logit-Normal, or even U-shaped priors) operate blindly, failing to adapt to this heterogeneity. They waste computational budget on linear, low-curvature segments while under-fitting highly curved turns.

\begin{figure*}[t]
 \centering
 \includegraphics[width=0.95\linewidth]{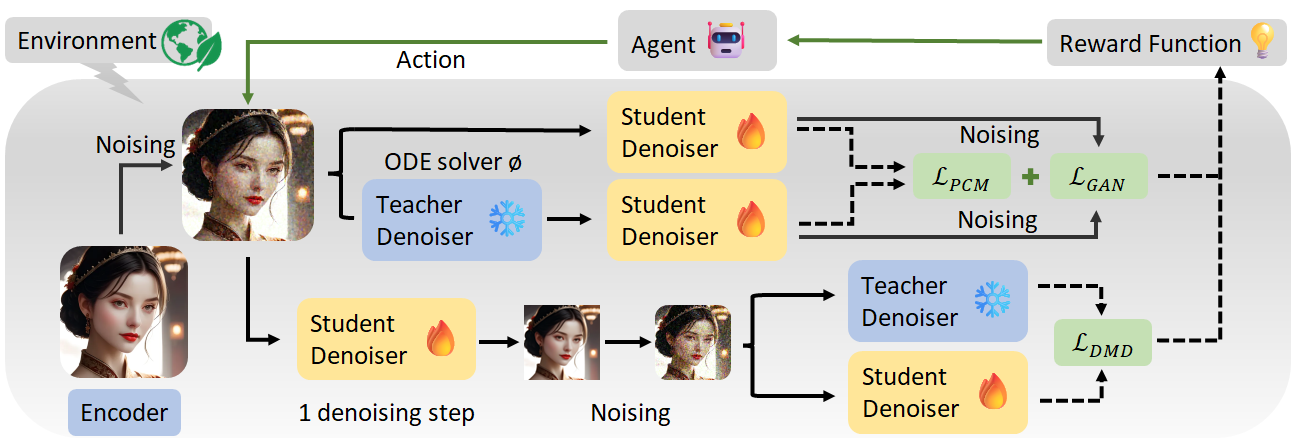}
 \caption{Training paradigm of \method.}
 \label{fig:sdxl_framework}
\end{figure*}

To address this, we propose \textbf{\method} (Fig.~\ref{fig:sdxl_framework}), a framework that reformulates consistency distillation as a curvature-adaptive decision process. We employ an RL agent as a probe, using consistency error as a curvature proxy to dynamically detect high-yield trajectory segments and prioritize them during training.

\subsection{RL as a Geometric Probe}
We formulate the sub-trajectory selection as a Markov Decision Process (MDP). Unlike prior works that use RL for content alignment (\eg, optimizing aesthetic scores), our agent optimizes the \textit{training dynamics} itself. 

\textbf{Semantic Stage Abstraction:}
A key challenge in applying RL to trajectory optimization is the potential explosion of the state space. To resolve this, we introduce a semantic stage abstraction. By partitioning the continuous trajectory $t \in [0, 1]$ into a small number of coarse geometric stages (\eg, $M=4$), we map the process into distinct semantic phases of generation: \textit{Initialization, Structural Formation, Texture Filling, and Final Refinement}.

\begin{itemize}[leftmargin=*]
    \item \textbf{State Space $\mathcal{S}$ (Relative Difficulty Profile):}
    To make informed decisions, the agent must perceive the relative difficulty across stages. We define the state $s_t$ as the ordinal ranking of the consistency losses {$\{ \mathcal{L}^{(1)}, \dots, \mathcal{L}^{(M)}\}$} associated with each sub-trajectory.
    \textit{Why this works:} By limiting $M=4$, the state space size is $|\mathcal{S}| = 4! = 24$. This compact representation transforms an intractable continuous control problem into a highly sample-efficient tabular Q-learning task, converging rapidly without overhead.

    \item \textbf{Action Space $\mathcal{A}$ (Curvature Focus):}
    At each step, the agent selects an action $a_t \in \{1, \dots, M\}$ for the next training iteration. This enables the model to allocate resources directly to the identified bottleneck.

    \item \textbf{Reward Function $\mathcal{R}$ (Baseline Advantage):}
    Directly using the raw loss difference as a reward is problematic because different sub-trajectories exhibit inherent variations in optimization difficulty (\ie, some segments naturally have higher loss values). To address this, we adopt a baseline-based advantage strategy. We maintain an exponential moving average (EMA) of the consistency loss for each sub-trajectory $m$, denoted as $B_t(m)$. The reward $r_t$ is defined as the improvement of the current consistency loss $\mathcal{L}_{\text{con}}$ over its historical baseline, scaled by a factor $\lambda_r$:
\begin{equation}\label{eq:reward}
    r_t = \lambda_r \cdot (B_t(a_t) - \mathcal{L}_{\text{con}}),
\end{equation}
where the baseline is updated as $B_{t+1}(a_t) \leftarrow \beta B_t(a_t) + (1-\beta)\mathcal{L}_{\text{con}}$ with $\lambda_r=100$ and $\beta=0.95$. This formulation encourages the agent to select sub-trajectories that yield higher optimization benefits, effectively realizing a dynamic curriculum learning strategy.
\item \textbf{Policy Update:} We use tabular Q-learning rather than a neural policy. 
    After observing the next state $s_{t+1}$, the Q-table is updated with $\alpha=0.1$ and $\gamma=0.9$:
\begin{equation*}
    Q(s_t,a_t) \leftarrow Q(s_t,a_t) + \alpha \big[r_t + \gamma\, \textstyle{\max_{a}} Q(s_{t+1},a) - Q(s_t,a_t)\big],
\end{equation*}
We adopt an $\epsilon$-greedy exploration policy, with $\epsilon$ linearly decayed from $1.0$ to $0.1$ during the first 20,000 training steps. This state-dependent update enables temporal credit assignment beyond myopic loss-proportional sampling.
\end{itemize}

\subsection{Hybrid Objective for High-Fidelity Distillation}\label{sec:Hybrid_Objective}

While RL optimizes \textit{where} to learn, we must ensure the model learns \textit{effectively} at the chosen locations. Relying solely on the consistency loss is prone to mode collapse or blurry artifacts in few-step regimes. To address this, we integrate DMD and Adversarial Consistency into the objective.

\subsubsection{DMD.}
Inspired by \cite{yin2024one}, we incorporate a DMD objective to explicitly align the student's generated distribution with the teacher's target distribution. Specifically, we minimize the Kullback-Leibler (KL) divergence between the student's marginal distribution $\mathbb{P}_{\boldsymbol\theta}^{\text{student}}(\mathbf{x}_0)$ and the teacher's distribution $\mathbb{P}_{\boldsymbol\phi}^{\text{teacher}}(\mathbf{x}_0)$:
\begin{equation*}
\mathcal{L}^{\text{DMD}} = D_{\text{KL}}\big(\mathbb{P}_{\boldsymbol\theta}^{\text{student}}(\mathbf{x}_0) \,|\, \mathbb{P}_{\boldsymbol\phi}^{\text{teacher}}(\mathbf{x}_0)\big).
\end{equation*}
Following the derivation in \cite{yin2024one}, the gradient of this objective is approximated via the score difference evaluated at perturbed samples. Let $\mathbf y = \boldsymbol f_{\boldsymbol\theta}(\mathbf x_t, t)$ denote the clean sample predicted by the student. To provide a stable optimization signal, we introduce a re-noising step to compute the gradient update:
\begin{equation}
\nabla_{\theta}\mathcal{L}^{\text{DMD}} \approx \mathbb{E}\left[ \left(\boldsymbol s^{\text{teacher}}(\mathbf x_\tau) - \boldsymbol s^{\text{student}}(\mathbf x_\tau) \right) \nabla_{\theta} \boldsymbol f_{\boldsymbol\theta}(\mathbf x_t, t) \right],
\end{equation}
where $\tau \sim \mathcal{U}(0,1)$ is a re-noising timestep, $\boldsymbol \epsilon \sim \mathcal{N}(0, I)$, and $\mathbf x_\tau = (1-\tau)\mathbf y + \tau\boldsymbol\epsilon$ is the re-noised sample (using the Flow Matching noise schedule). Here $\boldsymbol s^{\text{teacher}}$ and $\boldsymbol s^{\text{student}}$ are the score functions evaluated at the noisy state $\mathbf x_\tau$.
Specifically, the computation involves three steps: 
1) The student predicts a clean sample $\mathbf y$.
2) $\mathbf y$ is re-noised to $\mathbf x_\tau$. 
3) The scores are computed at $\mathbf x_\tau$. 
For the student score $\boldsymbol s^{\text{student}}(\mathbf x_\tau)$, we efficiently approximate it using the student model itself via the vector field relationship, avoiding the need for an extra discriminator network.

\subsubsection{Score Estimation in FM.}
Applying the DMD loss requires evaluating the score function $\boldsymbol s_t(\mathbf  x) = \nabla_{\mathbf  x_t} \log p_t(\mathbf x_t)$. While traditional diffusion models predict the noise $\boldsymbol\epsilon$ to estimate the score, our FM teacher predicts the vector field  $\boldsymbol v_{\boldsymbol \theta}$. Under the optimal transport probability path $p_t(\mathbf x|\mathbf x_0) = \mathcal{N}((1-t)\mathbf x_0, t^2 \mathbf I)$, we leverage Tweedie's formula to analytically derive the score function directly from the velocity output: 
\(s_\theta(\mathbf x_t, t) = -[{\mathbf x_t + (1-t)\boldsymbol v_{\boldsymbol\theta}}]/{t}
\). 

This rigorous transformation allows us to seamlessly compute the score difference for the DMD loss using exclusively the velocity outputs. The complete step-by-step mathematical derivation is provided in Appendix~\ref{sec:appendix_score}.

\subsubsection{Adversarial Consistency Loss.}
To enhance distribution consistency in few-step generation, we introduce an adversarial loss defined as: 
\begin{equation*}
\mathcal L^{\mathrm{adv}}(\boldsymbol \theta, \boldsymbol \theta^{-}; \boldsymbol \phi, m) = \operatorname{ReLU}(1 + \mathcal D(\tilde{\mathbf x}_{s}, \boldsymbol c)) +\operatorname{ReLU}(1-\mathcal D(\hat{\mathbf x}_{s}, \boldsymbol c))
\end{equation*}
where \(\tilde{\mathbf x}_s = \boldsymbol f_{\boldsymbol \theta}^m(\mathbf x_{t_{n+1}}, t_{n+1}) +  \boldsymbol{\epsilon}_1\), \(\hat{\mathbf x}_s = \boldsymbol f_{\boldsymbol \theta^-}(\mathbf x_{t_n}^{\boldsymbol \phi}, t_n) +  \boldsymbol{\epsilon}_2\) with $\boldsymbol{\epsilon}_1$ and $\boldsymbol{\epsilon}_2$ denoting noise perturbations. Here $\mathcal D$ is a discriminator conditioned on prompts $\boldsymbol c$. 

This adversarial objective acts as a robust regularizer, ensuring that the consistency mapping strictly resides on the natural image manifold. Furthermore, this constraint performs a critical stabilizing function: as our DMD objective utilizes the student's own score estimates (\ie, self-distillation), the discriminator effectively prevents the model from drifting into degenerate solutions, counteracting the potential instability of bootstrapping. 

\subsection{Algorithm and Analysis of the Optimization Mechanism}\label{sec:Final loss function}
The training procedure of \textbf{\method} is summarized in Algorithm \ref{alg:cacfm} and illustrated in Fig.~\ref{fig:sdxl_framework}. Our framework operates in a dynamic closed loop: the RL agent perceives the geometric bottlenecks (State), selects the high-yield segment (Action), and the student model updates its flow via the hybrid loss (Consistency $+$ Adv $+$ DMD). This update alters the trajectory's geometry, reshaping the consistency error profile (Reward) and guiding the agent's next decision.

Crucially, to reduce variance from adversarial and distillation objectives, the reward $r_t$ \eqref{eq:reward} is evaluated solely based on the structural consistency loss $\mathcal{L}_{\text{PCM}}$, while the full hybrid loss $\mathcal{L}_{\text{total}}$ is used for back-propagation. Unless otherwise stated, we set $\lambda_{\mathrm{adv}}=0.1$, $\lambda_{\mathrm{DMD}}=0.5$, and the PCM consistency weight to $1.0$.

\begin{algorithm}[t]
\caption{Curvature-Adaptive Consistency Flow Matching (CACFM)}
\label{alg:cacfm}
\centering
\resizebox{0.97\textwidth}{!}{
\begin{minipage}{\textwidth}
    \begin{algorithmic}[1]
    \STATE \textbf{Input:} Data $\mathcal{D}$, Teacher $\boldsymbol{\phi}$, Student $\boldsymbol{\theta}$, Solver $\Psi$, Sub-trajectory partitions $\{s_j\}$.
    \STATE \textbf{Initialize:} Target ${\boldsymbol{\theta}}^- \leftarrow {\boldsymbol{\theta}}$, Q-table $Q(s,a)$.

    \REPEAT
    \STATE Sample batch $(\mathbf{z},\mathbf{c}) \sim \mathcal{D}$.
    \STATE \textbf{Observe Geometry:} Construct state $s_t$ by ranking current losses.
    \STATE \textbf{Select Curvature Focus:} Choose sub-trajectory $m \in \{1,\dots,M\}$ via $\epsilon$-greedy policy based on $Q(s_t, \cdot)$.
    \STATE Sample $t_{n+1}$ within segment $m$. Generate $\mathbf{x}_{t_{n+1}}$.
    \STATE Compute Teacher Target: $\mathbf{x}^\phi_{t_n} \leftarrow \text{SolverStep}(\Psi, \mathbf{x}_{t_{n+1}})$.
    \STATE Compute Student Prediction: $\tilde{\mathbf{x}}_{s_m} = \boldsymbol{f}_{\boldsymbol{\theta}}^m(\mathbf{x}_{t_{n+1}},t_{n+1})$.

    \STATE \textbf{Compute Hybrid Loss:} 
    \STATE $\mathcal{L}_{\text{total}} = \mathcal{L}^{\text{PCM}} + 0.1\mathcal{L}^{\mathrm{adv}} + 0.5\mathcal{L}^{\mathrm{DMD}}$

    \STATE \textbf{Update Model:} $\boldsymbol{\theta} \leftarrow \boldsymbol{\theta} - \eta\nabla_{\boldsymbol{\theta}}\mathcal{L}_{\text{total}}$.
    \STATE Update Target: ${\boldsymbol{\theta}}^- \leftarrow \mu{\boldsymbol{\theta}}^- + (1-\mu){\boldsymbol{\theta}}$.

    \STATE \textbf{Update Policy:} 
    \STATE Compute Reward $r_t$ \eqref{eq:reward}. Update $Q(s_t, m)$ via Q-learning.
    \UNTIL{convergence}
    \end{algorithmic}
\end{minipage}
}
\end{algorithm}

While prior works like \cite{fan2023sft} employed Policy Gradient methods primarily for \textit{alignment} (optimizing a specific reward model or discriminator score), our \method framework utilizes Q-learning for \textbf{structural discovery}. Our agent does not merely optimize the output; it probes the PF-ODE trajectory to identify and resolve geometric bottlenecks that hinder consistent distillation. 

By integrating reinforcement learning with the geometric properties of Flow Matching, our framework transcends standard acceleration techniques. The proposed approach yields three critical advantages:
\begin{itemize}[leftmargin=*, nosep]
    \item \textbf{Geometry-Awareness:} Unlike static strategies that function blindly, our agent can adapt to training characteristics, automatically allocating computational budget to high-error, high-curvature trajectory segments.
    
    \item \textbf{Emergent Curriculum:} The reward-guided optimization naturally induces a \textit{coarse-to-fine} training curriculum. Our empirical analysis reveals that the agent autonomously shifts its focus from global structure initialization at the trajectory boundaries to high-frequency detail refinement, mirroring human learning processes without manual scheduling.
    
    \item \textbf{Holistic Stability:} By combining the exploration capability of RL with the strict constraints of Adversarial and DMD losses, \method avoids the "reward hacking" often seen in RL, ensuring that the accelerated trajectory remains faithful to the teacher's original distribution.
\end{itemize}

\paragraph{Computational Efficiency.}
Crucially, the computational footprint introduced by our RL agent is negligible. Since the state space is low-dimensional (discrete loss rankings) and the Q-learning update is computationally lightweight, the additional cost is virtually non-existent compared to the gradient backpropagation through the massive Transformer backbones (\eg, FLUX with 12B parameters). Thus, \method achieves acceleration without incurring extra training latency.

\section{Experiments}
\subsection{Experimental Setup}
\paragraph{Dataset Configuration.}
We conduct our image generation experiments using the large-scale LAION dataset \cite{schuhmann2022laion} for model training. For evaluation purposes, we utilize a randomly selected 15,000-image subset from the CC3M dataset \cite{cc12m}, ensuring a representative sample for performance assessment.

\paragraph{Model Architecture.}
All text-to-image generation experiments are built upon FLUX \cite{flux2024} and Stable Diffusion XL (SDXL \cite{sdxl}) as our base architectures. This choice provides a robust foundation for comparing various acceleration techniques across different flow formulations (\ie, Rectified Flow vs. DDPM).

\paragraph{Evaluation Protocol.}
We employ a comprehensive suite of metrics to rigorously assess both generation quality and prompt adherence:
\begin{itemize}[nosep]
\item Image Quality: FID \cite{fid} for distribution similarity.
\item Human Preference: HPSv2 \cite{wu2023human} and PickScore \cite{kirstain2023pick}.
\item Aesthetic Quality: LAION Aesthetic Score \cite{schuhmann2022laion}.
\end{itemize}
 All metrics are uniformly computed on the 15K CC3M validation split to ensure strict, consistent, and fair benchmarking across all compared methods.

\paragraph{RL Hyperparameter Configuration.}

To strike an optimal balance between routing granularity and computational overhead, we partition the generative trajectory into $M=4$ discrete segments for the RL agent.  Consequently, the state space cardinality is limited to $|\mathcal{S}| = M! = 24$.
We validate this choice by varying $M$ in 
Section~\ref{sec:Ablation} 
and measure the computational overhead of the $M=4$ setting in Section~\ref{sec:Training Efficiency}. The results show that this compact state space enables rapid policy convergence with negligible overhead compared to backbone training.

\begin{figure*}[b]
    \centering
    \begin{minipage}[t]{0.55\linewidth}
    \vspace{0pt}
        \centering
        \parbox{0.2\linewidth}{\centering\scalebox{0.8}{\textbf{Hyper-SD}}}\parbox{0.2\linewidth}{\centering\scalebox{0.8}{ \textbf{Schnell}}}\parbox{0.2\linewidth}{\centering\scalebox{0.8}{ \textbf{TDD}}}\parbox{0.2\linewidth}{\centering\scalebox{0.8}{ \textbf{Turbo}}}\parbox{0.2\linewidth}{\centering\scalebox{0.8}{\textbf{Ours}}}
        \includegraphics[width=1\linewidth]{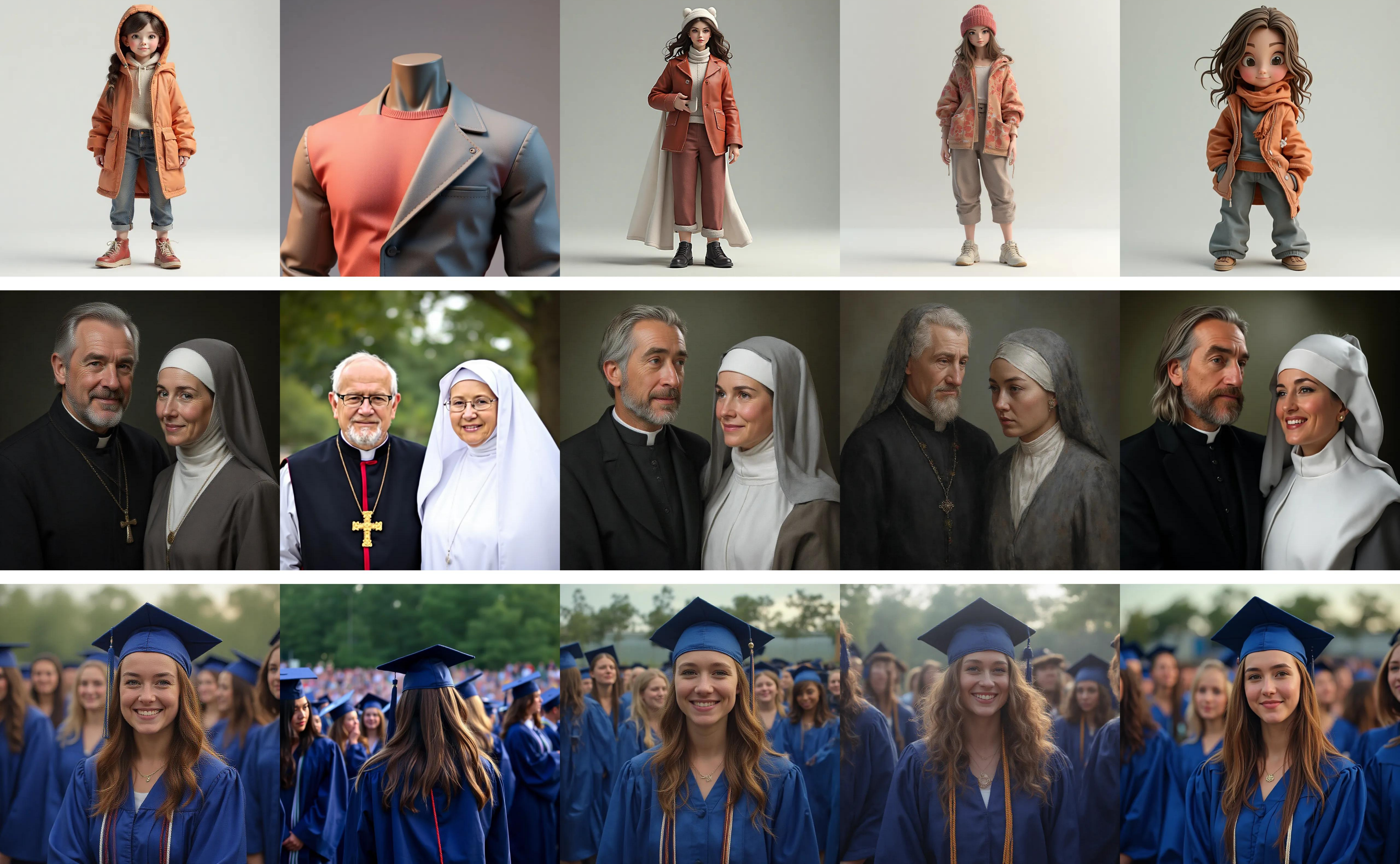}
        \caption{Qualitative comparisons with LoRA-based approaches on FLUX architecture.}
        \label{fig:flux_comp}
    \end{minipage}
    \hfill
\begin{minipage}[t]{0.42\linewidth}
    \centering
    \makeatletter\def\@captype{table}\makeatother
    \caption{FID Comparison on CC3M (FLUX). Best results are \textbf{bolded}.}
    \label{table:flux_multistep}
    \setlength{\tabcolsep}{0.8mm}
    \begin{tabular}{lccc}
        \toprule
        \multirow{2}{*}{\textbf{Methods}} & \multicolumn{3}{c}{\textbf{FID ($\downarrow$)}} \\
        \cmidrule(lr){2-4}
        & \scalebox{0.9}{4-Step} &  \scalebox{0.9}{8-Step} &  \scalebox{0.9}{16-Step} \\ \midrule
        \addlinespace[1.1ex]
        \scalebox{0.8}{Turbo \cite{sdturbo}}            & 58.22 & 46.24 & 43.66 \\
        \addlinespace[1.1ex]
        \scalebox{0.8}{Hyper-SD \cite{ren2024hyper}}    & 45.35 & 43.86 & 43.40 \\
        \addlinespace[1.1ex]
        \scalebox{0.8}{TDD \cite{wang2024target}}       & 45.81 & 41.65 & 41.12 \\
        \addlinespace[1.1ex]
        \scalebox{0.8}{Schnell \cite{flux2024}}         & 41.19 & 40.47 & 39.89 \\ 
        \addlinespace[1.1ex]
        \midrule
        \scalebox{0.8}{\method {(Ours)}}                   & \scalebox{0.95}{\textbf{39.19}} & \scalebox{0.95}{\textbf{36.96}} & \scalebox{0.95}{\textbf{37.62}} \\ \bottomrule
    \end{tabular}
\end{minipage}
\end{figure*}

\subsection{Comparative Analysis}
\paragraph{Baseline Methods. }
We compare \method against different SOTA approaches:
\begin{itemize}[leftmargin=*, nosep]
\item Consistency Models: LCM\cite{luo2023latent}, Hyper-SD \cite{ren2024hyper}, PCM\cite{wang2024phased}, TDD\cite{wang2024target}, TCD\cite{tcd}.
\item Flow-based Methods: InstaFlow (Insta. \cite{instaflow}).
\item Adversarial Distillation: Lightning (Light. \cite{sdxl-lightning}), Turbo \cite{sdturbo}, Schnell \cite{flux2024}.
\end{itemize}

\begin{figure*}[t]
\centering
\parbox{0.11\linewidth}{\centering \scalebox{0.8}{\textbf{Hyper-SD}}}\parbox{0.11\linewidth}{\centering\scalebox{0.8}{\textbf{LCM}}}\parbox{0.11\linewidth}{\centering\scalebox{0.8}{\textbf{Lightning}}}\parbox{0.11\linewidth}{\centering\small \scalebox{0.8}{\textbf{PCM}}}\parbox{0.11\linewidth}{\centering\scalebox{0.8}{\textbf{Instaflow}}}\parbox{0.11\linewidth}{\centering\scalebox{0.8}{\textbf{TCD}}}\parbox{0.11\linewidth}{\centering\scalebox{0.8}{\textbf{TDD}}}\parbox{0.11\linewidth}{\centering\scalebox{0.8}{\textbf{Turbo}}}\parbox{0.11\linewidth}{\centering\scalebox{0.8}{\textbf{Ours}}}
\includegraphics[width=0.99\linewidth]{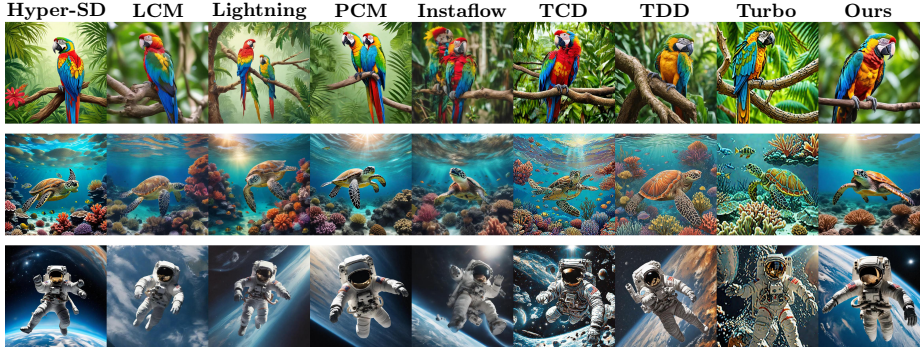}\\
\caption{Qualitative comparisons with LoRA-based approaches on SDXL architecture.}
\label{fig:sdxl_comp}
\end{figure*}

\paragraph{Qualitative Results. }
Figures~\ref{fig:flux_comp} and \ref{fig:sdxl_comp} present comprehensive visual comparisons. As illustrated, \method demonstrates superior capability in handling complex geometries. While baseline methods (\eg, Turbo, PCM) frequently suffer from structural collapse or severe texture blurring under few-step regimes, which are indicative of under-optimized high-curvature segments, \method maintains:
\begin{itemize}[leftmargin=*, nosep]
\item \textbf{Structural Integrity:} Significantly lower structural artifacts, suggesting the RL agent successfully reinforced the learning of difficult boundary conditions.
\item \textbf{High-Frequency Detail:} Better preservation of textures, attributed to the emergent curriculum that prioritizes refinement in later training stages.
\end{itemize}

\paragraph{Quantitative Results. }

~

\textbf{1) FID Performance:}
Tables~\ref{table:flux_multistep} and \ref{table:sdxl_multistep} 
present the FID scores evaluated on the CC3M dataset. Our \method achieves state-of-the-art results across all configurations. Notably, the performance gap is most pronounced in the extreme low-step regime (4-Step), where \method outperforms FLUX-schnell by a substantial margin of over 2 FID points. This empirical evidence confirms our hypothesis: when the inference budget is constrained, curvature-adaptive training is essential to resolve geometric bottlenecks that uniform sampling misses.
\begin{itemize}[leftmargin=*, nosep]
\item Consistency Model Comparison: \method outperforms LCM, PCM, and TCD, validating the benefit of dynamic trajectory optimization.
\item {SOTA Benchmarking}: Our approach rivals or surpasses Adversarial Distillation (Turbo) and Rectified-Flow (Insta.) methods, setting a new efficiency standard.
\end{itemize}

\begin{table}[t]
\caption{FID Comparison on CC3M (SDXL). Best results are \textbf{Bold}.}
\label{table:sdxl_multistep}
\centering
\small
\setlength{\tabcolsep}{0.6mm}
\resizebox{\linewidth}{!}{
\begin{tabular}{cccccccccc}
\toprule
\multirow{2}{*}{\textbf{Steps}}
& \multicolumn{9}{c}{\textbf{FID ($\downarrow$) Comparison across Different Methods} } \\ 
\cmidrule(lr){2-10}

 & Light.~\cite{sdxl-lightning} & Turbo~\cite{sdturbo} & LCM~\cite{luo2023latent} & Hyper-SD~\cite{ren2024hyper} & PCM~\cite{wang2024phased} & Insta.~\cite{instaflow} & TDD~\cite{wang2024target} & TCD~\cite{tcd} & \textbf{Ours} \\ \midrule

4  & 37.49 & 52.90 & 45.57 & 39.43 & 37.26 & 38.13 & 41.75 & 46.40 & \textbf{35.29} \\
8  & 38.28 & 65.25 & 43.67 & 41.63 & 39.30 & 35.60 & 46.00 & 49.51 & \textbf{34.42} \\
16 & 40.22 & 77.13 & 43.33 & 44.12 & 40.47 & 34.43 & 51.22 & 54.68 & \textbf{33.49} \\ 
\bottomrule
\end{tabular}
}
\end{table}

\vspace{0.5mm}
\textbf{2) Human-Centric Evaluation:} Tables~\ref{tab:aes-flux} and \ref{tab:aes-xl} report metrics assessing perceptual quality and text alignment.
\begin{itemize}[leftmargin=*, nosep]
\item Consistent Performance: \method achieves top-tier performance (consistently ranking 1st or 2nd) across the vast majority of configurations.
\item Robustness: Our approach maintains stable improvements across increasing steps. Crucially, the high Aesthetic and PickScores suggest that the emergent curriculum effectively transitions focus to fine-grained details (Phase 3) in later training stages, directly boosting perceptual fidelity. 
\end{itemize}

\begin{table*}[t]
\caption{Aesthetic evaluation on FLUX. Best results are \textbf{Bold}.}\label{tab:aes-flux}
\centering
\resizebox{1.0\textwidth}{!}{
\setlength{\tabcolsep}{1mm}
\begin{tabular}{lcccccccccccc}
\toprule
\multirow{2}{*}{\textbf{Methods}} & \multicolumn{3}{c}{Step 4} & \multicolumn{3}{c}{Step 8} & \multicolumn{3}{c}{Step 16} \\
\cmidrule(lr){2-4} \cmidrule(lr){5-7} \cmidrule(lr){8-10}
 & HPS & Aesthetic & PickScore & HPS & Aesthetic & PickScore & HPS & Aesthetic & PickScore \\
\midrule
Turbo & 0.1293 & 5.5666 & 17.1633 & 0.1372 & 5.7197 & 17.0438 & 0.2658 & 5.7665 & 21.3205 \\
Hyper-SD & 0.2626 & 5.6734 & 21.2479 & \textbf{0.2747} & 5.7120 & 21.3677 & \textbf{0.2757} & 5.7617 & 21.4478 \\
TDD & 0.2454 & 5.6165 & 20.7101 & 0.2610 & 5.6576 & 20.9471 & 0.2609 & 5.6383 & 20.8932 \\
Schnell & \textbf{0.2713} & 5.5204 & 21.3134 & 0.2696 & 5.4914 & 21.2159 & 0.2671 & 5.4802 & 21.1154 \\
\midrule
Ours$^{\text{(rank)}}$ & 0.2638$^{(2)}$ & \textbf{5.7955}$^{(1)}$ & \textbf{21.3160}$^{(1)}$ & {0.2683}$^{(3)}$ & \textbf{5.7961}$^{(1)}$ & \textbf{21.3818}$^{(1)}$ & 0.2754$^{(2)}$ & \textbf{5.7967}$^{(1)}$ & \textbf{21.4520}$^{(1)}$ \\
\bottomrule
\end{tabular}
}
\end{table*}

\begin{table*}[t]
\caption{Aesthetic evaluation on SDXL: Our method consistently achieves top-tier performance, often ranking among the top three or even attaining the best results. Best results are \textbf{Bold}, and the superscript indicates the rank of our method.}\label{tab:aes-xl}
\centering
\resizebox{1.0\textwidth}{!}{
\setlength{\tabcolsep}{1mm}
\begin{tabular}{lccccccccc}
\toprule
\multirow{2}{*}{\textbf{Methods}} & \multicolumn{3}{c}{Step 4} & \multicolumn{3}{c}{Step 8} & \multicolumn{3}{c}{Step 16} \\
\cmidrule(lr){2-4} \cmidrule(lr){5-7} \cmidrule(lr){8-10}
 & HPS & Aesthetic & PickScore & HPS & Aesthetic & PickScore & HPS & Aesthetic & PickScore \\
\midrule
Lightning & 0.2666 & 5.7135 & \textbf{21.3174} & 0.2721 & 5.8287 & 21.2698 & 0.2660 & 5.8600 & 21.0609 \\
Turbo & 0.2587 & 5.3267 & 20.6089 & 0.2471 & 5.2256 & 20.2537 & 0.2393 & 5.1566 & 20.0273 \\
Hyper-SD & \textbf{0.2855} & 5.8806 & 21.2701 & 0.2862 & 5.9284 & \textbf{21.4551} & 0.2898 & 5.9372 & 21.4797 \\
LCM & 0.2431 & 5.4165 & 20.8978 & 0.2493 & 5.4680 & 20.9471 & 0.2473 & 5.4863 & 20.8295 \\
PCM & 0.2663 & 5.6441 & 21.0911 & 0.2731 & 5.7256 & 21.1061 & 0.2704 & 5.7573 & 20.9734 \\
InstaFlow & 0.2472 & 5.5360 & 21.0710 & 0.2522 & 5.5813 & 21.1527 & 0.2560 & 5.6136 & 21.1934 \\
TDD & 0.2609 & 5.7519 & 20.9910 & 0.2602 & 5.8571 & 20.8012 & 0.2511 & 5.8673 & 20.4932 \\
TCD & 0.2576 & 5.5966 & 20.7705 & 0.2543 & 5.6558 & 20.5408 & 0.2450 & 5.6267 & 20.2597 \\\midrule
Ours$^{\text{(rank)}}$ & 0.2764$^{(2)}$ & \textbf{5.8931}$^{(1)}$ & 21.2241$^{(3)}$ & \textbf{0.2875}$^{(1)}$ & \textbf{5.9423}$^{(1)}$ & 21.3321$^{(2)}$ & \textbf{0.2932}$^{(1)}$ & \textbf{5.9823}$^{(1)}$ & \textbf{21.5532}$^{(1)}$ \\
\bottomrule
\end{tabular}
}
\end{table*}

\paragraph{Inference Scalability. }
As presented in Tables~\ref{table:flux_multistep} and \ref{table:sdxl_multistep}, 
\method exhibits strong zero-shot scalability. Although the RL policy is trained with a fixed granularity of $M=4$ segments, it consistently improves generation quality when extrapolated to more inference steps ($N=8$ \& $16$). For instance, on the FLUX backbone, the FID score substantially improves from 39.19 (4-step) to 36.96 (8-step). 
This suggests that RL-guided optimization better regularizes the probability flow trajectory, allowing users to trade computation for perceptual quality at inference time. In contrast, conventional distillation baselines frequently overfit to their specific training step counts, failing to capitalize on increased inference budgets.

\begin{figure*}[t]
    \centering
    \includegraphics[height=0.292\textwidth]{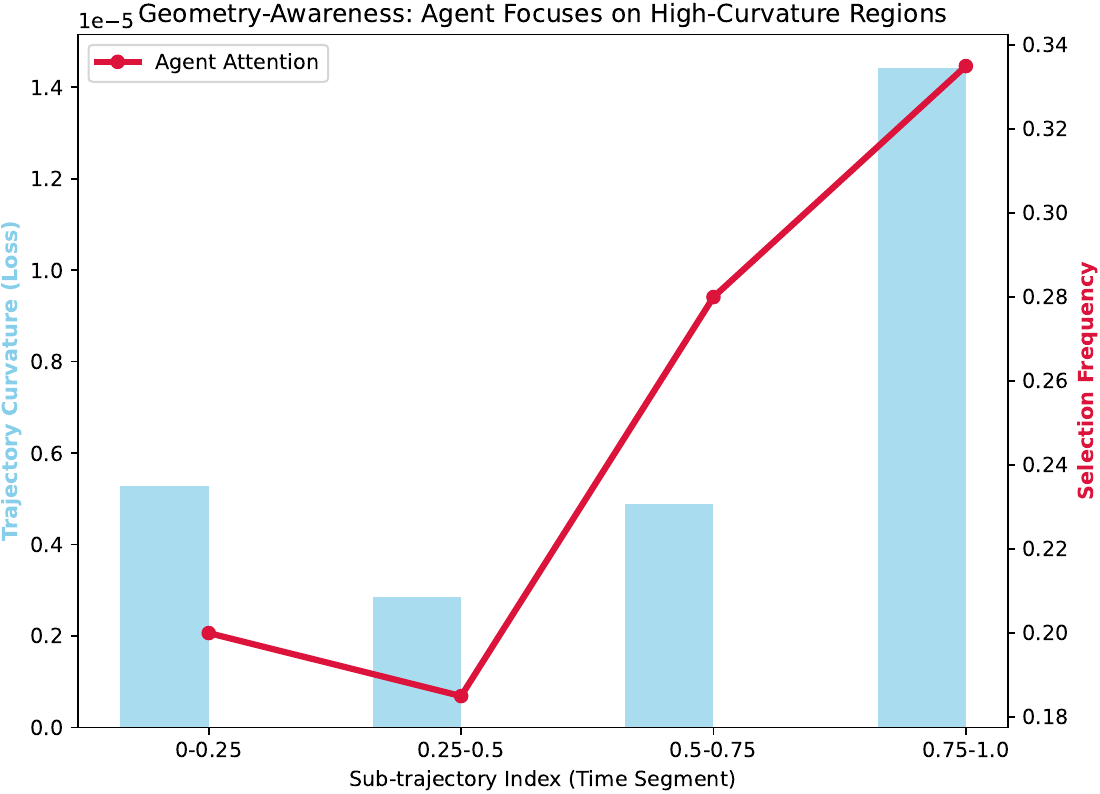}
    \includegraphics[height=0.3\textwidth]{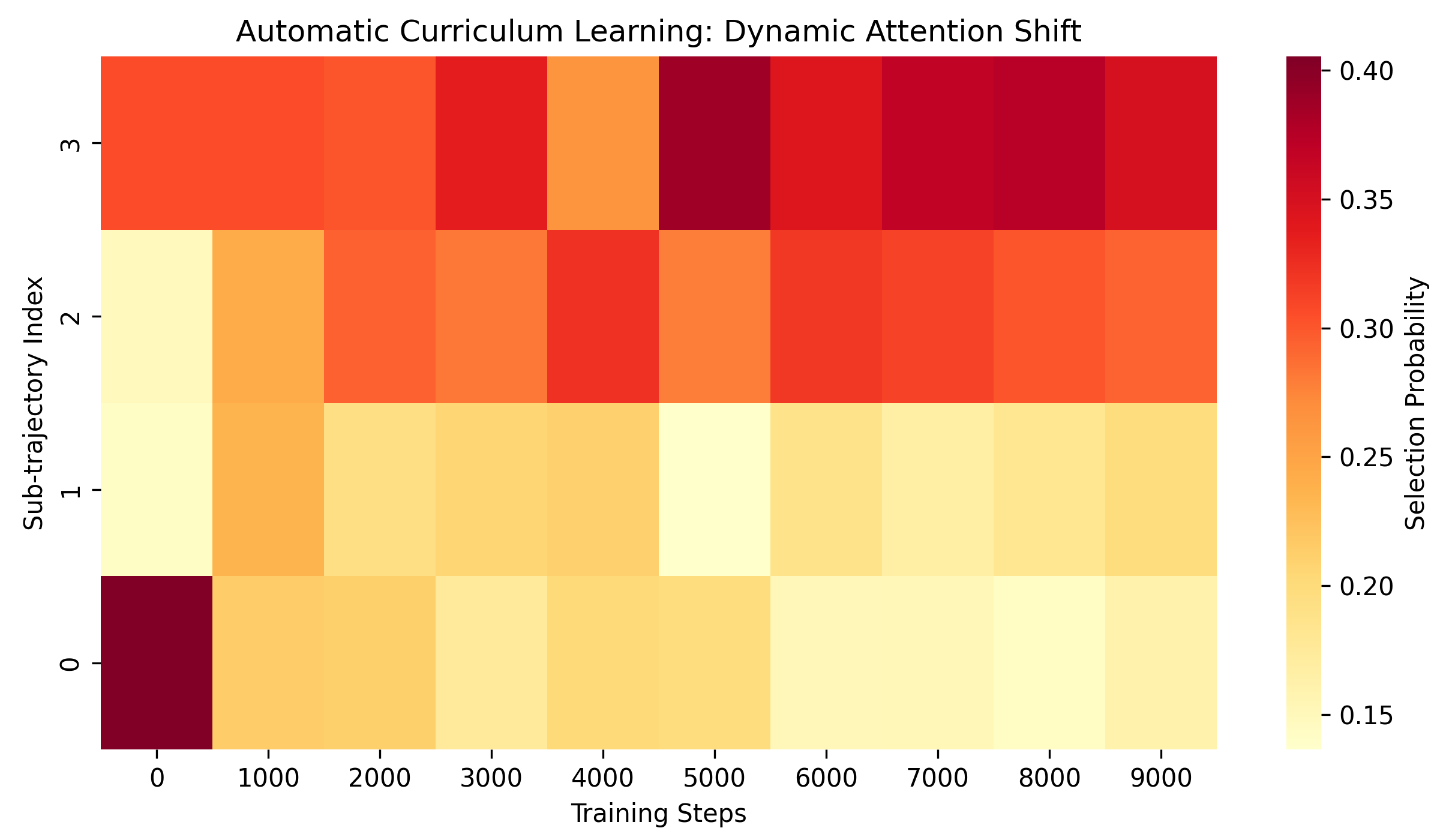}
    \caption{\textbf{Analysis of the Learned Policy.} \textit{Left:} The agent's selection frequency (Red) perfectly aligns with the U-shaped optimization difficulty profile (Blue Bars), validating its geometry-awareness. \textit{Right:} The selection probability heatmap shows an emergent curriculum, shifting focus from initialization to refinement (Phase 0 to 3) over time.}
    \label{fig:rl_analysis}
\end{figure*}

\subsection{Analysis: Uncovering Geometric Heterogeneity}
\label{sec:analysis}

This section presents the experimental results that systematically validate the core observations outlined at the beginning of the paper.

\vspace{2mm}
\textbf{U-Shaped difficulty profile. }
We computed the \textit{Oracle Consistency Error} (proxy for ground truth difficulty) across the trajectory using a converged SDXL teacher model.
Fig.~\ref{fig:rl_analysis} (Left) reveals a profound geometric insight:
\begin{itemize}[leftmargin=*, nosep]
    \item The ground truth error follows a {"U-shaped" profile}, indicating that the \textit{boundary conditions} initialization ($t \to 0$) and final refinement ($t \to 1$) are the actual geometric bottlenecks.
    \item Our RL agent's learned policy (Red Line) perfectly correlates with this ground truth ($\rho > 0.95$), automatically shifting focus to the boundaries.
\end{itemize}
The widely used Logit-Normal prior is geometrically misaligned for consistency distillation. \method outperforms it by simply "learning where to look." 

\vspace{2mm}
\textbf{Emergent Curriculum Learning. }
Fig.~\ref{fig:rl_analysis} (Right) visualizes the temporal evolution of the agent's policy. We observe an emergent coarse-to-fine curriculum:
\begin{itemize}[leftmargin=*, nosep]
    \item \textit{Early Training (0--20k steps):} The agent prioritizes the $t \to 0$ phase (Phase 0), focusing heavily on global structural formation.
    \item \textit{Late Training (20k+ steps):} The focus autonomously shifts towards $t \to 1$ (Phase 3), dedicating capacity to refining high-frequency details and textures.
\end{itemize}
Crucially, this sophisticated behavior emerges purely from reward-driven optimization, entirely without any manual heuristic scheduling.

\subsection{Training Efficiency and Wall-Clock Time Analysis.}\label{sec:Training Efficiency}
A potential concern regarding the introduction of the RL agent and discriminator is the computational overhead per training iteration. To evaluate the true training efficiency, we compare \method against a strong baseline, PCM \cite{wang2024phased}, under a strict \textit{fixed wall-clock time} budget, rather than a fixed number of iterations. This comparison accounts for the additional forward/backward passes introduced by our auxiliary modules. 
We conducted training on 8 accelerators with 80GB device memory each. 
As reported in Table~\ref{tab:time_efficiency}, we measured the FID on the CC3M validation set at different training durations (12h, 24h, and 36h). 

\textbf{Results and Discussion:} 
The empirical results in Table~\ref{tab:time_efficiency} demonstrate that \method consistently maintains a superior FID-to-time ratio compared to PCM across all checkpoints. Notably, \method achieves an FID of 40.82 within just 24 hours, effectively surpassing the performance of PCM at 36 hours (FID 41.65). 

While \method introduces a marginal computational overhead of approximately 18\% per iteration due to the RL agent and DMD loss updates, the resulting gains in \textit{data efficiency} are substantial. The RL agent effectively acts as a dynamic curriculum learner, prioritizing high-error sub-trajectories and avoiding redundant training on easy segments. This accelerates the overall convergence rate, making \method more time-efficient in practical training scenarios despite the increased model complexity.

\begin{table}[t]
    \centering
    \caption{\textbf{Wall-Clock Time Efficiency Comparison.} We compare the FID scores of PCM and our \method at fixed training intervals. Although \method incurs a slight overhead per iteration ($\sim$1.18$\times$ relative time per step), the RL-guided trajectory selection accelerates convergence significantly.}
    \label{tab:time_efficiency}
    \small
    \resizebox{0.8\columnwidth}{!}{
    \setlength{\tabcolsep}{3mm}
    \begin{tabular}{lcccc}
    \toprule
     \multirow{2}{*}{\textbf{Methods}} &  \multirow{2}{*}{\makecell[c]{\textbf{Relative Time}\\\textbf{per Step}}} & \multicolumn{3}{c}{\textbf{FID ($\downarrow$) on CC3M over Time}} \\
    \cmidrule(lr){3-5} 
    & & \textbf{12 Hours} & \textbf{24 Hours} & \textbf{36 Hours} \\
    \midrule
    PCM  & 1.00$\times$ & 46.52 & 43.10 & 41.65 \\
    \method (Ours) & 1.18$\times$ & \textbf{44.15} & \textbf{40.82} & \textbf{39.19} \\
    \bottomrule
    \end{tabular}}
\end{table}

\subsection{Ablation Study}\label{sec:Ablation}

\begin{table}[t]
\centering
\caption{Ablation Study: adversarial vs. DMD loss. Avg Rank ($\downarrow$) denotes the mean ranking across all 9 evaluation dimensions.}\label{tab:aes-ablation}
\resizebox{0.99\textwidth}{!}{
\setlength{\tabcolsep}{1.1mm}
\small
\begin{tabular}{lcccccccccc}
\toprule
\multirow{2}{*}{\textbf{Methods}} & \multicolumn{3}{c}{Step 4} & \multicolumn{3}{c}{Step 8} & \multicolumn{3}{c}{Step 16} & \multirow{2}{*}{\multirow{2}{*}{\makecell[c]{\textbf{Avg}\\\textbf{Rank} ($\downarrow$)}} } \\
\cmidrule(lr){2-4} \cmidrule(lr){5-7} \cmidrule(lr){8-10}
& HPS & Aesthetic & PickScore & HPS & Aesthetic & PickScore & HPS & Aesthetic & PickScore & \\
\midrule
Naive CFM \scalebox{0.8}{(Uniform)} & 0.236 & 5.432 & 20.69 & 0.264 & 5.602 & 21.12 & 0.272 & 5.646 & 21.16 & 4.11 \\
CFM Logit-Normal    & 0.232 & 5.458 & 20.70 & 0.265 & 5.598 & 20.21 & 0.270 & 5.637 & 21.11 & 4.89 \\
CFM Loss-Aware      & 0.236 & 5.436 & 20.77 & 0.263 & 5.588 & 21.16 & 0.270 & 5.659 & 21.17 & 4.00 \\
\method w/o DMD     & 0.239 & 5.485 & 20.78 & 0.263 & 5.594 & 21.16 & 0.270 & 5.623 & 21.18 & 3.78 \\
\method w/o RL      & 0.235 & 5.430 & 20.71 & 0.265 & 5.610 & 21.20 & 0.271 & 5.658 & 21.23 & 3.22 \\
\method \scalebox{0.8}{(Ours)}      & \textbf{0.276} & \textbf{5.893} & \textbf{21.22} & \textbf{0.288} & \textbf{5.942} & \textbf{21.33} & \textbf{0.293} & \textbf{5.982} & \textbf{21.55} & \textbf{1.00} \\
\bottomrule
\end{tabular}}
\end{table}

\paragraph{Static vs. Adaptive Geometric Priors. }
To isolate the benefit of adaptive RL, we compare \method against the following static sampling strategies in Table~\ref{tab:aes-ablation}.
\begin{itemize}[nosep]
    \item \textbf{Naive CFM (Uniform)} and \textbf{Logit-Normal CFM} serve as baselines for fixed geometric assumptions.
    \item \textbf{CFM Loss-Aware} is a dynamic loss-proportional sampler that updates stage probabilities using the EMA of per-stage consistency losses; it lacks the state transition modeling and temporal credit assignment of Q-learning. 
\end{itemize}
\begin{table*}[t]

\begin{minipage}[t]{0.52\textwidth} 
\vspace{0pt}
\centering
\caption{Comparison of dynamic adaptive sampling strategies on FLUX 4-step generation. Loss-Aware and MAB adapt online but treat stages myopically, while CACFM uses Q-learning over ranking states.}
\label{tab:adaptive_sampler}
\resizebox{1\columnwidth}{!}{
\setlength{\tabcolsep}{1mm}
\small
\begin{tabular}{lccc}
\toprule
\textbf{Sampling Strategy} & \textbf{HPS} & \textbf{Aesthetic} & \textbf{PickScore} \\
\midrule
Loss-Aware (EMA) & 0.236 & 5.436 & 20.77 \\
 MAB ($\epsilon$-greedy) & 0.242 & 5.515 & 20.84 \\
\midrule
\textbf{CACFM (Ours)} & \textbf{0.276} & \textbf{5.893} & \textbf{21.22} \\
\bottomrule
\end{tabular}}
\end{minipage}
\hfill
\begin{minipage}[t]{0.44\textwidth} 
\vspace{0pt}
\centering
\caption{Hyperparameter ablation on the number of semantic stages $M$ under the 4-step setting. $M=4$ achieves the best FID by balancing trajectory granularity and exploration cost.
}
\label{tab:m_sensitivity}
\small
\resizebox{\columnwidth}{!}{
\setlength{\tabcolsep}{3.2mm}
\begin{tabular}{lcc}
\toprule
\textbf{Stages} & \textbf{State Space} & \textbf{FID} ($\downarrow$) \\
\midrule
$M=3$ & $3!=6$ & 39.85 \\
$M=4$ & $4!=24$ & \textbf{39.19} \\
$M=6$ & $6!=720$ & 39.51 \\
\bottomrule
\end{tabular}}
\end{minipage}
\vspace{0mm}
\end{table*}

\noindent
The results show that \method outperforms both uniform and heuristic-based methods. Notably, CFM Logit-Normal (which prioritizes intermediate steps) underperforms relative to Naive CFM, further validating our finding that the true difficulty profile is U-shaped, not bell-shaped. Compared with dynamic but myopic samplers such as Loss-Aware and MAB (Table~\ref{tab:adaptive_sampler}), CACFM further benefits from state-dependent long-horizon scheduling.

\paragraph{Impact of Hybrid Loss. }
We evaluate the contribution of the DMD loss component. As shown in the "w/o DMD loss" rows, removing distribution alignment leads to a noticeable drop in aesthetic scores. This indicates that while RL effectively optimizes the structural consistency (fidelity), the DMD loss is crucial for aligning the student's fine-grained texture distribution with the teacher's, validating the synergy of our hybrid objective.

\paragraph{Sensitivity to the Number of Semantic Stages. }
The number of semantic stages $M$ determines the granularity at which the flow trajectory is ranked and routed by the Q-table policy. A smaller $M$ gives a compact state space and fast exploration, but it can merge intervals whose consistency losses follow different dynamics. A larger $M$ provides a finer ordering of local bottlenecks, but the induced permutation state space grows factorially and can make early exploration less reliable. 

We therefore validate this design choice by varying the number of semantic stages while keeping the 4-step sampling setting fixed. As shown in Table~\ref{tab:m_sensitivity}, $M=4$ achieves the best balance between routing resolution and exploration cost: $M=3$ is too coarse and merges distinct bottlenecks, whereas $M=6$ enlarges the state space to $720$ states and slows early Q-table exploration. This trend supports using a moderate discretization of the trajectory rather than treating more stages as monotonically better.

\section{Conclusions and Limitations}\label{sec:limit}
\paragraph{Conclusions. } 
We presented \method, a framework that reformulates consistency distillation as geometric trajectory optimization. By employing an RL agent to dynamically target high-curvature bottlenecks and integrating a unified Flow-DMD objective, \method induces an emergent training curriculum that moves beyond static sampling priors. This adaptive strategy improves trajectory learning, mitigates structural artifacts, and achieves state-of-the-art few-step generation on large-scale FLUX and SDXL backbones.

\paragraph{Limitations. } 
As a distillation approach, \method's generative quality is inherently bounded by the teacher model's vector field manifold and by the expressiveness of the student model used to approximate it. Moreover, although adaptive trajectory routing improves few-step generation, extreme compression to single-step inference remains challenging. Highly curved or topologically complex flows cannot always be faithfully approximated by a single linear transition, which may lead to minor artifacts or loss of fine-grained details in difficult samples.

\section*{Acknowledgements}
We thank the members of Huawei Large Model Data Technology Lab for their valuable suggestions and support throughout this work. The authors also benefited from the academic environment and exchanges at the Department of Statistics and Data Science, Tsinghua University, and are grateful to the faculty members and fellow students for their guidance, insights, and helpful discussions. We also thank the anonymous reviewers and area chairs for their insightful and constructive comments, which helped improve the quality of this work.

\bibliographystyle{splncs04}
\bibliography{main}

\clearpage
\appendix
\renewcommand{\theHsection}{appendix.\arabic{section}}

\begin{center}
    \Large
    \textbf{{C}urvature-{A}daptive {C}onsistency {F}low {M}atching: Autonomous Trajectory Optimization via Reinforcement Learning}\\[0.5em]
    Supplementary Material\\[1.0em]
\end{center}

\section{Derivation of the Score Function from Vector Field}
\label{sec:appendix_score}

This section provides a detailed mathematical derivation for the \textbf{Score Estimation in FM} part of Section~\ref{sec:Hybrid_Objective} in the main text, establishing the exact relationship between the predicted vector field $v_\theta$ and the score function $s_\theta(x_t,t)$ under the Flow Matching framework.

Given the optimal transport probability path \(p_t(x_t \mid x_0)=\mathcal{N}((1-t)x_0,t^2I)\), Tweedie's formula expresses the score function in terms of the posterior expectation of the clean data \(\hat{x}_0\):
\begin{equation*}
    s(x_t, t) = -[x_t - \mu_t(\hat{x}_0)]/{\sigma_t^2} = -[x_t - (1-t)\hat{x}_0]/{t^2}
\end{equation*}

In the Flow Matching formulation, the forward trajectory is defined via linear interpolation as $x_t = (1-t)x_0 + t\epsilon$. Taking the derivative with respect to time $t$ yields the velocity (vector field):
\begin{equation*}
    v = {\mathrm{d} x_t}/{\mathrm{d}t} = \epsilon - x_0 \implies \epsilon = v + x_0
\end{equation*}

By substituting $\epsilon$ back into the trajectory equation, we can express the clean data $\hat{x}_0$ purely algebraically in terms of the current state $x_t$ and the predicted velocity $v_\theta$:
\begin{align*}
    x_t &= (1-t)\hat{x}_0 + t(v_\theta + \hat{x}_0); \\
    x_t &= \hat{x}_0 + t v_\theta, \quad
    \hat{x}_0 = x_t - t v_\theta.
\end{align*}

Finally, substituting this expression for $\hat{x}_0$ back into Tweedie's formula, we obtain the exact score function for Flow Matching:
\begin{align*}
    s_\theta(x_t, t) &= -\frac{x_t - (1-t)(x_t - t v_\theta)}{t^2} = -\frac{x_t - (x_t - t v_\theta - t x_t + t^2 v_\theta)}{t^2}\\
                     &= -\frac{t x_t + t v_\theta - t^2 v_\theta}{t^2} = -\frac{x_t + (1-t)v_\theta}{t}.
\end{align*}

This derivation confirms that the DMD loss can be rigorously applied to velocity-predicting Flow Matching models without theoretical discrepancy.

\section{Discussion about implement of state space}
 For computational tractability, $s_t$ may be encoded as: 
A \textbf{Lehmer code} (bijection to integers $\{0, 1, \dots, M!-1\} )$, or 
A \textbf{one-hot vector} $\in \{0,1\}^{M!}$ , or 
A \textbf{pairwise comparison matrix} $C \in \{0,1\}^{M \times M}$ , where \( C_{ij} = \mathbb{I}(L^{(i)} < L^{(j)}) \). 
We use the sub-trajectory loss defined in Eq.~\eqref{eq:pcm_loss} of the main text, and define the reward loss in Section~\ref{sec:Final loss function} of the main text.

\section{Convergence Analysis of Reinforcement Learning}

 \begin{figure}[t!]
 \centering
 \includegraphics[width=0.9\linewidth]{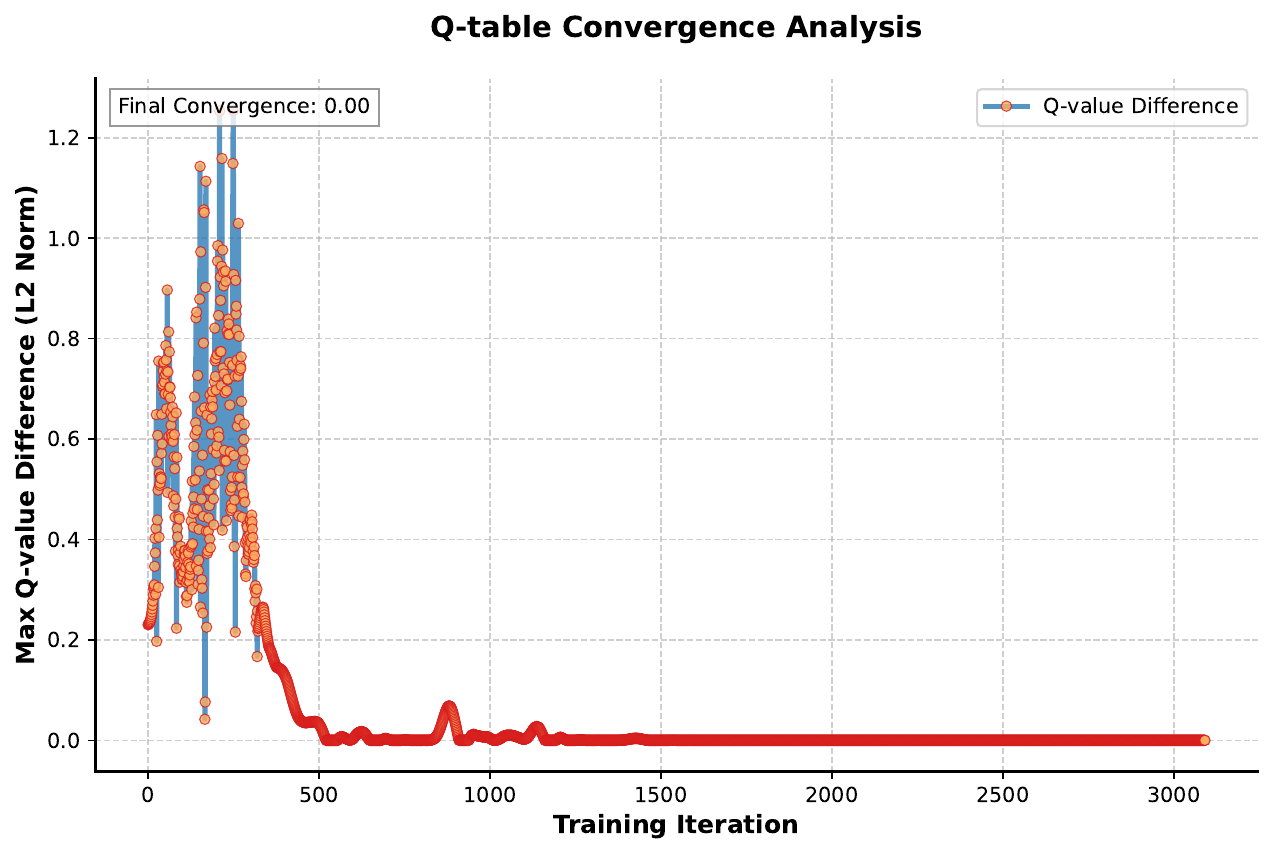}
\caption{Convergence of max Q-value differences across training iterations. The x-axis represents training epochs after window smoothing, while the y-axis shows the Q-value differences. The monotonic decrease of the $\ell_2$-norm demonstrates asymptotic stability.}
 \label{fig:q_table}
\end{figure}

\paragraph{Policy Convergence (Optimal Action Stability).}
 For each state (represented by a row in the Q-table), the index of the maximum Q-value remains invariant during the final training iterations. This stability indicates that the learned policy $\pi^*(s) = \arg \max_a Q(s,a)$ has reached a stationary solution.
 Specifically, the global average stable ratio of Q-table is $95.8\%$.

\paragraph{Value Function Convergence.}
The maximum Q-value per state demonstrates asymptotic stability, as evidenced by the convergence of the squared $\ell_2$-norm of temporal differences: 
\begin{equation}\label{eq:convergen of q_table}
 \sum_{s \in \mathcal{S}} \Big( \max_a Q_{t+1}(s,a) - \max_a Q_t(s,a) \Big)^2 \to 0 
\end{equation}
In Fig.~\ref{fig:q_table}, we plot the left hand side of (\ref{eq:convergen of q_table}) with increasing steps.
The empirical results presented in Fig.~\ref{fig:q_table} confirm this theoretical expectation.

\paragraph{Optimal Strategy Dominance.} 
 The optimality gap satisfies the following condition for all states: 
\[
\min_{s \in \mathcal{S}} \Big( \max_a Q(s,a) - \max_{a' \neq a^*} Q(s,a') \Big) \geq \epsilon > 0 
\] 
where $\epsilon$ constitutes a problem-dependent positive constant. This guarantees the distinguishability of optimal actions from suboptimal alternatives. 
Specifically, in the last several iterations of our experiments, the average gap is exceeds 0.08.

\end{document}